\documentclass[sigconf]{acmart}

\usepackage{multirow}
\usepackage{color, xcolor}
\usepackage{soul}
\usepackage{colortbl}
\usepackage{balance}
\usepackage{enumitem}

\newcommand{\p}{\colorbox[rgb]{0.9976624375240293, 0.9116955017301038, 0.8710495963091118}}

\newcommand{\ppppp}{\colorbox[rgb]{0.9879430988081508, 0.5608612072279893, 0.43537101114955795}}
\newcommand{\e}{\colorbox[rgb]{0.9101883890811227, 0.9469127258746636, 0.9812995001922338}}
\newcommand{\ee}{\colorbox[rgb]{0.8495501730103806, 0.9075432525951557, 0.9616147635524798}}
\newcommand{\eee}{\colorbox[rgb]{0.7904959630911188, 0.8681737793156478, 0.9419300269127259}}
\newcommand{\eeee}{\colorbox[rgb]{0.7063437139561707, 0.8290196078431372, 0.9127104959630911}}
\newcommand{\eeeee}{\colorbox[rgb]{0.6047058823529413, 0.7839753940792004, 0.8791387927720108}}

\AtBeginDocument{%
  \providecommand\BibTeX{{%
    \normalfont B\kern-0.5em{\scshape i\kern-0.25em b}\kern-0.8em\TeX}}}

\copyrightyear{2021} 
\acmYear{2021} 
\setcopyright{acmlicensed}
\acmConference[CIKM '21]{Proceedings of the 30th ACM International Conference on Information and Knowledge Management}{November 1--5, 2021}{Virtual Event, QLD, Australia}
\acmBooktitle{Proceedings of the 30th ACM International Conference on Information and Knowledge Management (CIKM '21), November 1--5, 2021, Virtual Event, QLD, Australia}
\acmPrice{15.00}
\acmDOI{10.1145/3459637.3482440}
\acmISBN{978-1-4503-8446-9/21/11}

\acmSubmissionID{2030}

\begin{document}
\fancyhead{}

\title[Integrating Pattern- and Fact-based Fake News Detection via Model Preference Learning]{Integrating Pattern- and Fact-based Fake News Detection via Model Preference Learning}

\author{Qiang Sheng, Xueyao Zhang}
\affiliation{
  \institution{Institute of Computing Technology,\\ Chinese Academy of Sciences}
  \institution{University of Chinese Academy of Sciences}
}
\email{shengqiang18z@ict.ac.cn}
\email{zhangxueyao19s@ict.ac.cn}

\author{Juan Cao}
\affiliation{%
  \institution{Institute of Computing Technology, \\ Chinese Academy of Sciences}
  \institution{University of Chinese Academy of Sciences}
}
\email{caojuan@ict.ac.cn}

\author{Lei Zhong}
\affiliation{%
  \institution{Institute of Computing Technology, \\ Chinese Academy of Sciences}
\institution{University of Chinese Academy of Sciences}
}
\email{zhonglei18s@ict.ac.cn}

\thanks{The authors are at the Key Lab of Intelligent Information Processing of Chinese Academy of Sciences. Qiang Sheng and Xueyao Zhang contributed equally. \\ Corresponding Author: Juan Cao.}
\renewcommand{\shortauthors}{Sheng, et al.}

\begin{abstract}
To defend against fake news, researchers have developed various methods based on texts. These methods can be grouped as 1) \emph{pattern-based} methods, which focus on shared patterns among fake news posts rather than the claim itself; and 2) \emph{fact-based} methods, which retrieve from external sources to verify the claim's veracity without considering patterns. The two groups of methods, which have different preferences of textual clues, actually play complementary roles in detecting fake news. However, few works consider their integration.
In this paper, we study the problem of integrating pattern- and fact-based models into one framework via modeling their preference differences, i.e., making the pattern- and fact-based models focus on respective preferred parts in a post and mitigate interference from non-preferred parts as possible.
To this end, we build a Preference-aware Fake News Detection Framework (\textbf{Pref-FEND}), which learns the respective preferences of pattern- and fact-based models for joint detection. We first design a heterogeneous dynamic graph convolutional network to generate the respective preference maps, and then use these maps to guide the joint learning of pattern- and fact-based models for final prediction.
Experiments on two real-world datasets show that Pref-FEND effectively captures model preferences and improves the performance of models based on patterns, facts, or both. 
\end{abstract}

\ccsdesc[500]{Information systems~Data mining}
\ccsdesc[500]{Computing methodologies~Natural language processing}

\keywords{fake news detection, preference learning, graph neural networks, pattern mining, fact-checking}

\maketitle

\begin{figure}[ht]
\vspace{-0.5cm}
\setlength{\abovecaptionskip}{0cm}
\setlength{\belowcaptionskip}{-0.5cm}
	\centering
	\includegraphics[width=\linewidth]{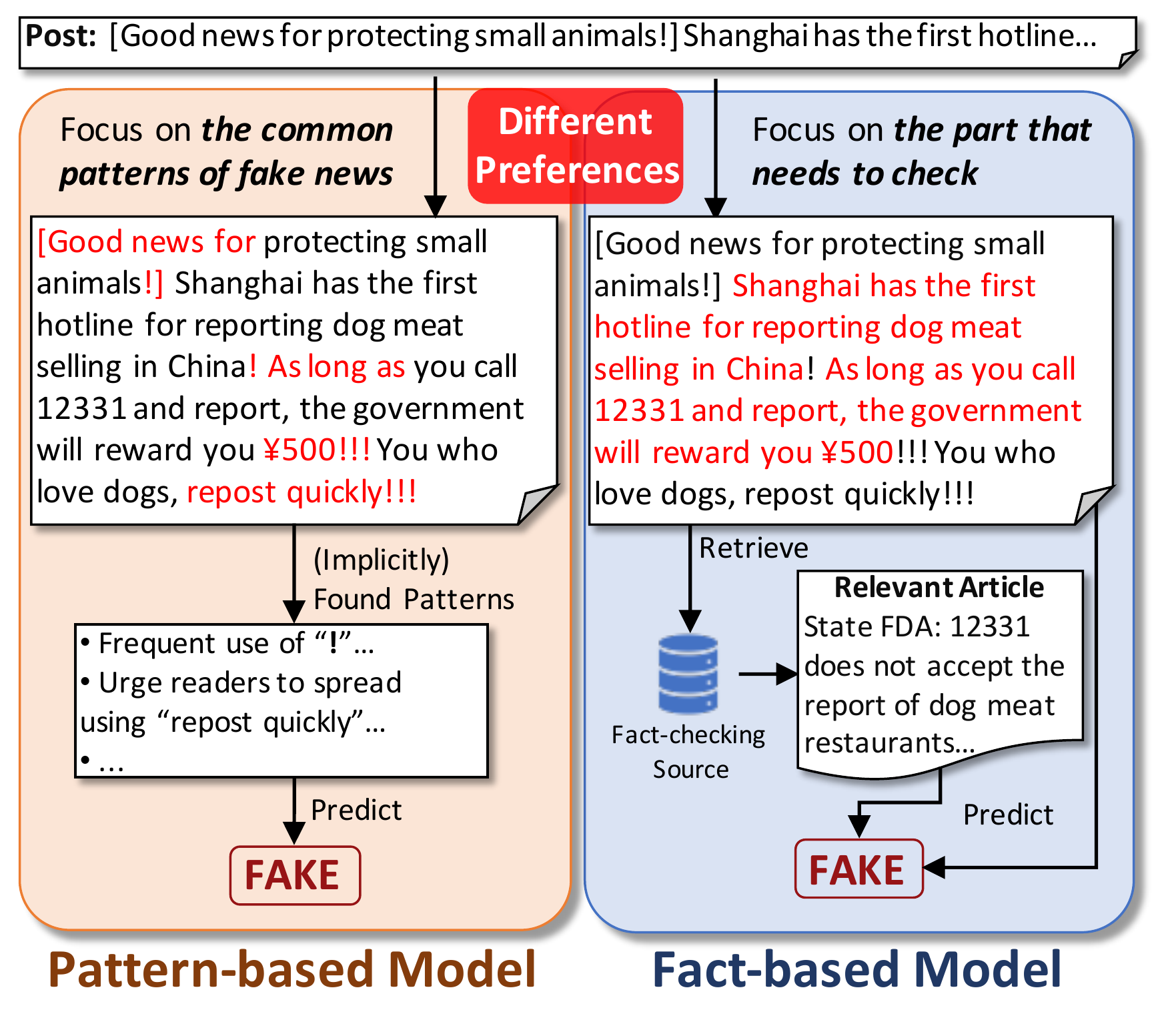}
	\caption{A motivating example. Ideally, given the same news post, the pattern-based and the fact-based model have \textit{different preferences} on textual clues to predict whether the post is fake. The post is translated into English.}
	\label{fig:motivating_eg}
\end{figure}

\section{Introduction}
\label{sec:intro}
Fake news that spreads on ``online'' social media continually causes ``offline'' real-world harms in crucial domains, such as politics~\citep{fisher2016pizzagate}, finance~\citep{kogan2019fake}, and public security~\citep{Bangladesh-lynchings}. The most recent example is the COVID-19 infodemic~\citep{infodemic} where thousands of fake news pieces spread through social media~\citep{wiki-covid-19}. Under such severe circumstances, developing fake news detection systems has been critical for maintaining a trustful online news ecosystem.

To detect fake news on social media, researchers propose to extract hand-crafted features or deep-learning features~\citep{cao-survey} from contents, social contexts, propagation networks, etc. In this paper, we focus on the deep learning methods based on textual contents, which can be grouped as: \textbf{1) Pattern-based methods} (e.g., \citep{eann, hdsf, fakeflow, dual-emotion}), which aim at learning shared features (patterns) among fake news posts and expect these features to generalize to unseen news posts. Once trained, they can operate without reliance on external resources. \textbf{2) Fact-based methods} (e.g., \citep{declare, majing-claim, wulianwei-aaai, nguyen-eacl}), which focus on the claim's veracity itself with the help from external fact-checking sources. The key difference between these two methods lies in their different preferences of textual clues. As Figure~\ref{fig:motivating_eg} shows, given the post about a newly opened hotline that accepts reports of dog meat selling, an ideal pattern-based model tends to predict the veracity relying more on the highly frequent use of exclamation marks or the words that urge readers to repost (``repost quickly''), while an ideal fact-based one retrieves to check whether the hotline accepts reports of dog meat selling. From the motivating example, we see that the different preferences of the two models lead to their \textit{complementary} roles. This inspires us to integrate patten- and fact-based models with considering their preferences, which may bring additional gain for fake news detection. However, how to effectively integrate them remains under-explored by existing works.

In this paper, we first study the problem of integrating the pattern- and fact-based models into one framework. The challenge lies in preference modeling: The models, though having different preferences, generally lack the constraints to make themselves focus on preferred parts and ignore non-preferred parts of inputs. As a consequence, a pattern-based model may overfit by \textit{memorizing} frequently shown non-preferred words (e.g., event-specific words) in the training set, and a fact-based one may be distracted from the part that describes a verifiable event. Moreover, the preference of each model should be dynamically determined with contexts, making rule-based modeling inapplicable.

\begin{figure*}[t]
\setlength{\belowcaptionskip}{-0.2cm}
	\centering
	\includegraphics[width=\textwidth]{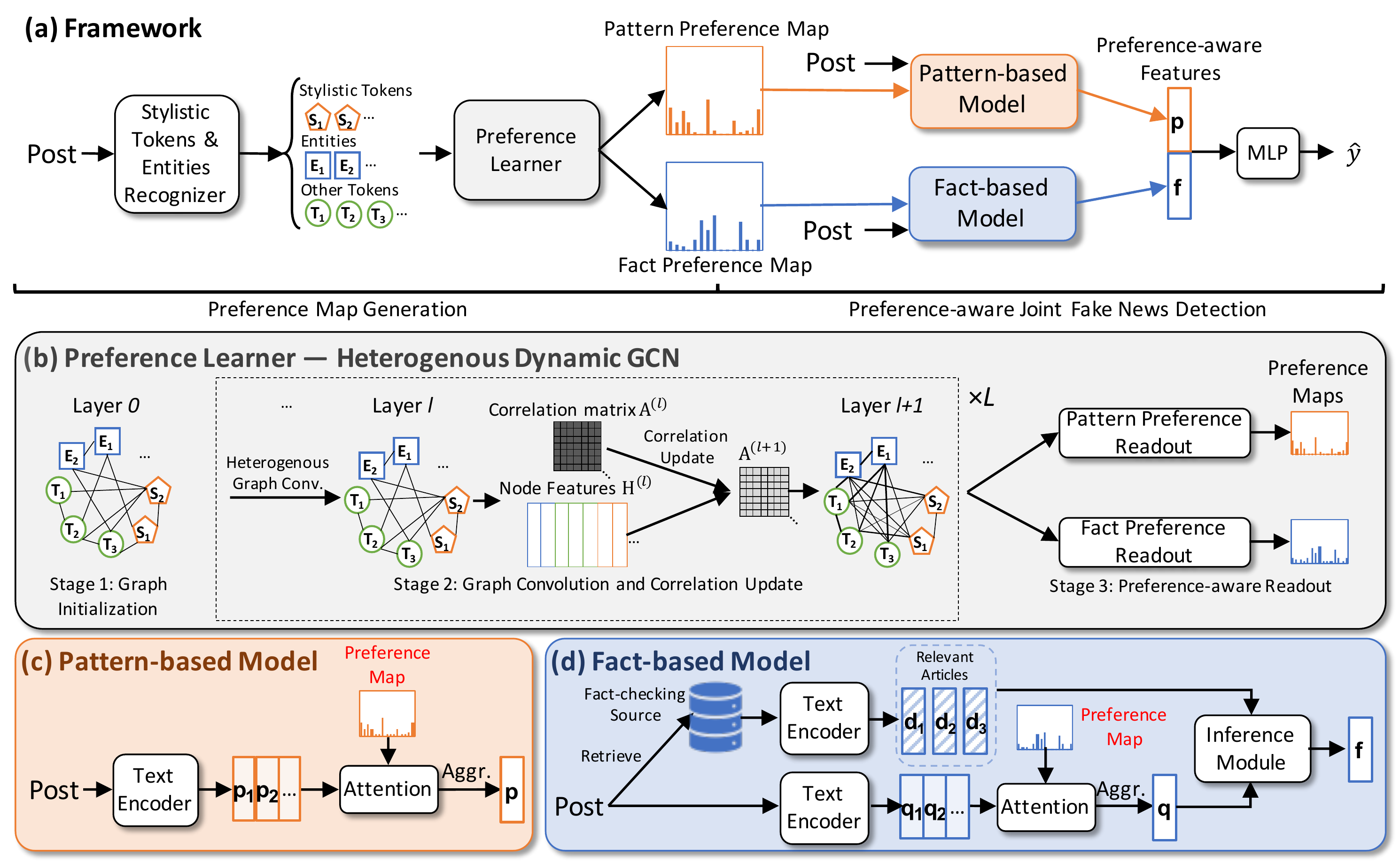}
	\caption{Architecture of Pref-FEND. (a) Overall framework. The post is divided into three sets: stylistic tokens, entities, and others. Then they are fed into a preference learner to generate two preference maps, which highlight the preferred information of the downstream detection models. The preference-aware features are concatenated for final prediction. (b) The Preference Learner, heterogeneous Dynamic GCN, leverages a heterogeneous graph convolution to aggregate multi-type neighbors and updates the correlation matrix every layer. Two readouts use the correlation matrix of the $L$-th layer to generate preference maps. Only parts of nodes and edges are shown. (c) and (d) exemplify how a pattern-based and a fact-based model works with the preference map, respectively. With the Maps, (c) and (d) attend to helpful tokens for capturing patterns or facts.}
	\label{fig:arch}
\end{figure*}

To address these aforementioned challenges, we propose to learn the models' preferences simultaneously with joint fake news detection and build Preference-aware Fake News Detection Framework (\textbf{Pref-FEND}). As Figure~\ref{fig:arch}(a) shows, Pref-FEND generates preference maps to assist each model to focus on its expected preferred part. Specifically, we exploit the prior knowledge verified by existing works (e.g., \cite{castillo, dual-emotion, fever}) to recognize cue tokens for patterns and facts, and obtain three sets of tokens (i.e., stylistic tokens, entities, and others). Then, we use a graph-based preference learner to dynamically learn the preferences within the contexts, as presented in Figure~\ref{fig:arch}(b). We construct a heterogeneous graph using these sets and design a Heterogeneous Dynamic Graph Convolutional Network (HetDGCN) for node correlation learning. The final correlation matrix is used by two preference-aware readout functions to generate the Fact and the Pattern Preference Map, respectively. For joint fake news detection, we feed the post and the Map to each model and fuse their last-layer features for final prediction. During training, besides the normal classification loss, we design two auxiliary losses as enhancements, which respectively minimize the similarity between the two maps and the classification loss when the input maps are exchanged and ground-truth labels are reversed. Experimental results on two real-world datasets show that our proposed Pref-FEND can effectively learn the models' preferences and improve the performance of both single preference (pattern- or fact-based) and integrated (pattern-and-fact-based) models.

Our contributions are summarized as follows:
\begin{itemize}[leftmargin=10pt]
	\item To the best of our knowledge, our work is the first that combines pattern- and fact-based fake news detection. We discuss their complementary roles in fake news detection and propose to consider their preferences for better integration.
	\item We propose a novel framework, Pref-FEND, which leverages a heterogeneous dynamic GCN to learn model preferences and effectively integrates them for fake news detection.
	\item Extensive experiments on two newly constructed datasets demonstrate the effectiveness of Pref-FEND on learning models' preferences and improving the detection performance for both single-preference models and integrated models. The code and datasets are available at \url{https://github.com/ICTMCG/Pref-FEND}.
\end{itemize}

\section{Related Work}
\subsection{Fake News Detection}
Fake news detection aims at automatically classifying a news piece as real or fake. Existing methods mostly capture features from contents (texts or/and images) and social contexts that generate in the spread process, such as propagation networks~\citep{pattern-driven, shukai-propagation, fang, Propagation2Vec}, user profiles~\citep{shukai-user-profile}, metadata~\citep{sadhan}, and crowd feedbacks~\citep{varying-shades, defend, dual-emotion}. In this paper, we focus on the text-based methods which can be grouped as:

\textit{Pattern-based Fake News Detection.} As fake news often contains opinionated and inflammatory language to attract readers~\citep{shukai-survey}, common patterns that are different from those in real news are shared across fake news pieces of different topics. In the very first work on evaluating information credibility on social media, \citet{castillo} list a series of post-based features, including the length, whether the post contains exclamation or question marks, etc. Following this line, \citet{Volkova} injects subjectivity, psycholinguistic, and moral foundations features into deep neural networks (CNNs and RNNs). \citet{style-fake-news} focuses on writing styles. Some works attempt to differentiate the patterns across multiple topical categories~\citep{domain-diff, mdfend}. A recent trend of pattern-based methods is to refocus on the sentiment and emotional patterns~\citep{ajao, giachanou, dual-emotion, fakeflow}, as the use of eye-catching terms in deceptive and fake posts may manipulate the readers' emotions~\citep{stop-clickbait}.

\textit{Fact-based Fake News Detection.} These methods judge the veracity of a news piece more objectively, with references to pre-constructed external resources such as knowledge graphs~\citep{MKEMN, DETERRENT} and online encyclopedias~\citep{fever}. A more flexible way is to directly use articles retrieved by search engines as evidence to predict the news veracity~\citep{declare, multifc}. \citet{declare} use post-specific attention to model the post-article interactions, while the following works~\citep{majing-claim, wulianwei-ijcai, wulianwei-aaai, nguyen-eacl} consider text entailment, such as coherence and conflicts using the attention mechanism.

Note that the claims provided by the datasets for evaluation of fact-based methods is generally \textit{normalized} by the human fact-checkers to be declarative and concise, so they are not suitable to evaluate the pattern-based ones. In this paper, we construct two new datasets (in English and Chinese) by referring to existing datasets and external sources for evaluation of pattern-and-fact-based methods.

Different from the above methods, our work do not develop better pattern- or fact-based methods, but integrate the existing ones for comprehensively detecting fake news based on texts.

\subsection{Graph Neural Networks for Text Mining} Due to its expressive power for integrating structural and semantic information, graph neural networks (GNNs) have been widely used for applications in text mining such as information extraction~\citep{re-graph} and sentiment analysis~\citep{absa-graph}. Most works use homogenous GNNs which treat nodes as the same type. \citet{heteGAT} leverages a heterogeneous GNN to handle multiple types of nodes such as topics and entities for text classification. Similarly, we use heterogeneous GNN to obtain the preference scores of each token, but our graph is dynamic as its node correlation matrix is adjustable (inspired by~\citep{eccv-dgcn}). The final adjusted correlations will be aggregated to obtain preference scores.

\section{Problem Statement}
Let $P$ be a news post on social media containing $n$ tokens. Let $D$ be the set of relevant articles of $P$. $D$ is retrieved from a fact-checking source $\mathcal{D}$. Following most existing works, we treat fake news detection as a binary classification problem. The ground-truth label $y$ is 1 if $P$ is fake, otherwise 0. We formulate the following tasks:

\textbf{Pattern-based Fake News Detection:} Given $P$, learn a function $f_P:f_P(P)\rightarrow \hat{y}$, such that it maximizes the predictive accuracy w.r.t. $y$.

\textbf{Fact-based Fake News Detection:} Given $P$, retrieve relevant articles $D$ from $\mathcal{D}$, learn a function $f_F:f_F(P, D) \rightarrow \hat{y}$, such that it maximizes the predictive accuracy w.r.t. $y$.

\textbf{Joint Pattern-and-Fact-based Fake News Detection:} Given $P$, $D$, a pattern-based model $f_P$ and a fact-based model $f_F$, learn a function $f:f(P, D, f_P, f_F) \rightarrow \hat{y}$, such that it maximizes the predictive accuracy w.r.t. $y$.

\section{Proposed Framework}
Figure~\ref{fig:arch}(a) overviews the architecture of the proposed Pref-FEND, whose goal is to learn the models' preferences and employ them for better joint fake news detection. Given a post $P$, Pref-FEND first respectively generates preference maps (i.e., token-level preference scores) for the pattern- and fact-based model with a heterogeneous dynamic GCN. Then, the preference maps are fed into the corresponding model along with $P$ to help the model focus on its preferred information. Finally, the models' output features are fused to predict if $P$ is real or fake. Besides the normal classification loss, we design two auxiliary losses as enhancements, whose goals are to minimize the similarity between the two maps and to minimize the classification loss when the input maps are exchanged and ground-truth labels are reversed, respectively. (see Section~\ref{sec:joint-det})

\subsection{Preference Map Generation}
Assuming that $P$ has $n$ tokens, a preference map is a score distribution of length $n$ where the $i$-th score represents to what extent the $i$-th token is preferred by the corresponding fake news detection model. For the pattern- and the fact-based model, we respectively generate Pattern Preference Map and Fact Preference Map
\begin{equation}
	\mathrm{\mathbf{m_P}}=[\mathrm{m_{P}}_i]_{i=1}^n,
	\mathrm{\mathbf{m_F}}=[\mathrm{m_{F}}_i]_{i=1}^n,
\end{equation}
where all scores are in $[0,1]$ and the sum of each map is $1$.

\begin{table}[t]
\centering
\small
\caption{\label{table:style} Types of Stylistic Tokens and References.}
\begin{tabular}{p{0.25\linewidth}|p{0.28\linewidth}p{0.28\linewidth}}
\hline
\textbf{Type} & \textbf{For Weibo} & \textbf{For Twitter} \\
\hline
Negation Words & \multicolumn{2}{c}{\multirow{4}{*}{HowNet Bilingual Dictionary~\citep{hownet}}}\\ \cline{1-1}
Degree Words & \multicolumn{2}{c}{}\\ \cline{1-1}
Sentiment Words &\multicolumn{2}{c}{}\\ \cline{1-1}
Proposition Words & \multicolumn{2}{c}{}\\ \hline
Punctuations & \multicolumn{2}{c}{\multirow{2}{*}{\citep{dual-emotion}}}\\ \cline{1-1}
Pronouns & \multicolumn{2}{c}{}\\ \hline
Emoticons & \multicolumn{2}{c}{List of Emoticons~\citep{emoticon} \citep{dual-emotion}}\\ \hline
Emotional \ \ \ \ \ \ \ \ \ \ \ \ \  Ontologies & Affective Lexicon Ontology~\cite{dalian} & NEC Emotion Lexicon~\cite{NRC}\\
\hline
\end{tabular}
\end{table}

\subsubsection{Stylistic Tokens \& Entities Recognition.}
As illustrated in Section~\ref{sec:intro}, a pattern-based model focuses on common patterns (generally, writing styles) while a fact-based one focuses on verifiable objective claims. To guide the map generation, we exploit the prior knowledge with reference to the existing pattern- and fact-based works. Specifically, we recognize tokens that are likely to represent writing styles or key objective elements. To indicate patterns, we recognize a set of \emph{stylistic tokens} $S=\{s_1,\ldots,s_{n_s}\}$ (e.g., emotional words, pronouns, punctuations)~\citep{dual-emotion}; and to indicate facts, we extract the \emph{entities} $E=\{e_1,\ldots,e_{n_e}\}$ because a verifiable claim generally contains at least one entity~\citep{fever}. These indicating tokens are derived using pre-constructed dictionaries and public tools.
In detail, to recognize stylistic tokens, we follow~\citep{dual-emotion}, which summarizes diverse emotion-related features and other useful linguistic features to represent textual patterns, and then generate a stylistic token table for each dataset. The types and references are shown in Table~\ref{table:style}. A simple exact matching is performed to recognize the stylistic tokens in posts. To recognize the entities, we use two public tools: \texttt{LAC}~\citep{lac}\footnote{\url{https://github.com/baidu/lac/}} for Chinese and \texttt{TexSmart}~\citep{texsmart1,texsmart2}\footnote{\url{https://ai.tencent.com/ailab/nlp/texsmart/en/index.html}. We use the v0.2.0 (Large).} for English. 
The tokens excluded by $S$ and $E$ are in a set $T=\{t_1,\ldots,t_{n_t}\}$ where $n_t=n-n_s-n_e$.

\subsubsection{heterogeneous Dynamic GCN}
Although the stylistic tokens and entities recognized by general dictionaries or tools provide a good prior to what tokens might be preferred, directly using the recognition result for map generation is insufficient: First, the coverage is limited, leading the map to overlook some other preferred and useful tokens for detection models; Second, a token's preference score should be dynamically determined in its context (i.e., the post) rather than static rules.
To enable the information of different types of nodes to dynamically and sufficiently interact with each other, we design a graph-based preference learner, Heterogeneous Dynamic Graph Convolutional Network (HetDGCN). As shown in Figure~\ref{fig:arch}(b), we first construct a heterogeneous graph that contains multi-type nodes (tokens) with a learnable correlation matrix (i.e., adjacent matrix). Then, we leverage a heterogeneous graph convolution to enable message passing among different types of nodes. The final preference scores are obtained using the learned correlation matrix. The stages are as follows:

\textit{Graph Initialization.} Recall that we have divided the tokens in $P$ into three parts: stylistic tokens $S$, entities $E$, and others $T$. To preserve their different roles, we construct a heterogeneous graph $G$, where each node corresponds to a token in $S$, $E$, or $T$ and the weight of each edge represents the correlation between the connected tokens.
The node representation is initialized with the pre-trained language model (here, BERT~\citep{bert}), denoted as $\mathrm{\mathbf{H}}^{(0)} \in \mathbb{R}^{n\times d}$ where $d$ is the dimensionality of each node vector. Note that this matrix is stacked with the representation of $S$, $E$, and $T$, i.e., $\mathrm{\mathbf{H}}^{(0)}=[\mathrm{\mathbf{H}}_S^{(0)};\mathrm{\mathbf{H}}_E^{(0)};\mathrm{\mathbf{H}}_T^{(0)}]$.

The edge weights (correlations) are initialized with calculating the cosine similarity of token pairs~\cite{heteGAT} which is scaled to $[0,1]$:
\begin{equation}
	\mathrm{\mathbf{A}}^{(0)}(i,j) = \frac{\mathrm{\mathbf{h}}_{i}^{(0)}\cdot\mathrm{\mathbf{h}}_{j}^{(0)}}{2\Vert\mathrm{\mathbf{h}}_{i}^{(0)}\Vert\Vert\mathrm{\mathbf{h}}_{j}^{(0)}\Vert}+0.5,
\end{equation}
where $\mathrm{\mathbf{h}}_{i}^{(0)}$ and $\mathrm{\mathbf{h}}_{j}^{(0)}$ are the initial node features, and $\mathrm{\mathbf{A}}^{(0)}(i,j)\in [0,1]$ is the initial weight of the edge connecting the $i$-th and the $j$-th node. Following~\citep{gcn}, we define the normalized correlation matrix of the $l$-th layer $\mathrm{\mathbf{\hat{A}}}^{(l)}=(\mathrm{\mathbf{D}}^{(l)})^{-\frac{1}{2}}\mathrm{\mathbf{A}}^{(l)}(\mathrm{\mathbf{D}}^{(l)})^{-\frac{1}{2}}$. $\mathrm{\mathbf{D}}^{(l)}$ is the degree matrix of the $l$-th layer where $\mathrm{\mathbf{D}}^{(l)}(i,i)=\sum_j \mathrm{\mathbf{A}}^{(l)}(i,j)$.

\textit{Graph Convolution \& Correlation Update.} Different types of nodes describe different aspects of the given text which we expect to distinguish for preference learning. Therefore, instead of using standard graph convolution for node interaction~\citep{gcn}, we use a heterogeneous graph convolution~\citep{heteGAT}, which separately handle the neighbors of different types and then aggregate the interacted features. Further, we use a dynamic correlation matrix which is updated each layer according to the present node similarity and expect the final correlations (edge weights) could reflect the bias of the nodes in the context. In detail, the feature matrix of $(l+1)$-th layer is calculated with
\begin{equation}
	\mathrm{\mathbf{H}}^{(l+1)} = \mathrm{ReLU}\left(\sum_{\tau \in \mathcal{T}} \mathrm{\mathbf{\hat{A}}}^{(l)}_{\tau}\mathrm{\mathbf{H}}^{(l)}_{\tau}\mathrm{\mathbf{W}}^{(l)}_{\tau}\right),
	\label{eq:h_(l+1)}
\end{equation}
where $\mathrm{\mathbf{\hat{A}}}^{(l)}_{\tau}$ is the submatrix of the correlation matrix of the $l$-th layer $\mathrm{\mathbf{\hat{A}}}^{(l)}$ whose rows contain all the nodes and columns record their correlation with nodes of the type $\tau \in \{S,E,T\}$. $\mathrm{\mathbf{W}}^{(l)}_{\tau}$ is the learnable weight matrix of the type $\tau$ in this layer.
Then, the correlation matrix is updated using

\begin{equation}
		\Delta\mathrm{\mathbf{A}}^{(l+1)} = \sigma\left(\mathrm{\mathbf{H}}^{(l+1)}\mathrm{\mathbf{W}}_A^{(l+1)}\mathrm{\mathbf{H}}^{(l+1)T}\right),
\end{equation}
\begin{equation}
	\mathrm{\mathbf{A}}^{(l+1)} = \alpha \mathrm{\mathbf{A}}^{(l)} + (1-\alpha) \Delta\mathrm{\mathbf{A}}^{(l+1)},
\end{equation}
where $\mathrm{\mathbf{W}}^{(l+1)}_{A}$ is the learnable weight matrix for updating correlations, $\sigma$ denotes the $\mathrm{sigmoid}$ function and $\alpha$ is a trade-off factor in $[0,1]$. 

\textit{Preference-aware Readout.} After the $L$-layer HetDGCN, we obtain the correlation matrix $\mathrm{\mathbf{A}}^{(L)}$, on which we expect to estimate the preference levels to pattern- and fact-based models of each token. For the $i$-th node, the pattern preference score $\mathrm{m_{P}}_i$ is calculated by its correlation with any nodes except those representing entity tokens:
\begin{equation}
	\mathrm{m_{P}}_i=\sum_{j=1}^{n}\mathrm{\mathbf{A}}^{(L)}(i,j) - \sum_{k=1}^{n_e}\mathrm{\mathbf{A}}^{(L)}_E(i,k).
\end{equation}

Similarly, the fact preference score excludes the correlation with the stylistic nodes:
\begin{equation}
	\mathrm{m_{F}}_i=\sum_{j=1}^{n}\mathrm{\mathbf{A}}^{(L)}(i,j) - \sum_{k=1}^{n_s}\mathrm{\mathbf{A}}^{(L)}_S(i,k).
\end{equation}

Finally, the preference maps are obtained by normalized the correlation sums of each token:
\begin{equation}
\mathrm{\mathbf{m_{P}}} = \bigg[\frac{\mathrm{m_{P}}_i}{\sum_j{\mathrm{m_{P}}_j}}\bigg]_{i=1}^n,
	\mathrm{\mathbf{m_{F}}} = \bigg[\frac{\mathrm{m_{F}}_i}{\sum_j{\mathrm{m_{F}}_j}}\bigg]_{i=1}^n.
\end{equation}

\subsection{Preference-aware Joint Fake News Detection}
\label{sec:joint-det}
As the fact-based and pattern-based models are diverse, we here use the typical pattern- and fact-based detection process to illustrate how to integrate the generated preference maps into them. Most specific models can be easily reformulated similarly to accommodate our framework.
 
\subsubsection{Pattern-based Model}
As shown in Figure~\ref{fig:arch}(c), a typical pattern-based uses a textual feature extractor to obtain a vector for final prediction. Here, we use the Pattern Preference Map as attention weights to make the model attend to its preferred tokens in the post $P$. For example, if the extractor is a BERT~\citep{bert} or an LSTM~\citep{lstm} whose output is $[\mathrm{\mathbf{p}}_1;\ldots;\mathrm{\mathbf{p}}_{n}]$, the aggregated vector is calculated as
\begin{equation}
	\mathrm{\mathbf{p}} = \sum_{i=1}^n \mathrm{m_{P}}_i\mathrm{\mathbf{p}}_i.
\end{equation}
Note that our preference map is at the token level, for the extractor that does not output $n$ vectors such as TextCNN~\citep{textcnn}, the map might be used before the extractor, right after we obtain token embeddings from pre-trained models.

\subsubsection{Fact-based Model}
In a typical fact-based model, the post $P$ are first used to retrieve from a fact-checking source to collect the related articles (or, evidence) $D$. Assuming $n_f$ articles are returned, we represent the articles in $D$ as $[\mathrm{\mathbf{d}}_1;\ldots;\mathrm{\mathbf{d}}_{n_f}]$. Then the post and evidence vectors are fed into an inference module, which is often designed to capture the complicated interactions such as coherence and conflicts between $P$ and $D$ (e.g., \cite{majing-claim, wulianwei-aaai}).
 The output vectors of inference module $\mathrm{\mathbf{f}}$, which implicitly represent the relationship of the post-evidence pairs, is used for final prediction.

To avoid the interference of non-check-worthy parts (e.g., the publisher's remark), the Fact Preference Map guides the inference module by using the attention mechanism to aggregate the token vectors in $P$ before post-evidence inference. The final vector is calculated as
\begin{equation}
	\mathrm{\mathbf{q}} = \sum_{i=1}^n \mathrm{m_{F}}_i\mathrm{\mathbf{q}}_i,
\end{equation}
\begin{equation}
	\mathrm{\mathbf{f}} = \mathrm{InferenceModule}(\mathrm{\mathbf{q}}, [\mathrm{\mathbf{d}}_1;\ldots;\mathrm{\mathbf{d}}_{n_f}]),
\end{equation}
where $\mathrm{\mathbf{q}}_i$ is the representation of the $i$-th token in $P$ for fact-based methods.

\subsubsection{Joint Detection.} 
For final prediction, we concatenate the output vectors of pattern- and fact-based models and feed it into a multi-layer Perceptron (MLP) and obtain the prediction $\hat{y}$:
\begin{equation}
	\hat{y}=\mathrm{MLP}([\mathrm{\mathbf{p}};\mathrm{\mathbf{f}}]).
\end{equation}

\subsubsection{Losses.} 
During training, we use three losses to supervise 1) the prediction of binary (fake and real) classification; and 2) the differentiation of the two preference maps. For the first goal, we minimize the cross-entropy loss between the prediction $\hat{y}$ and the label $y$ 
\begin{equation}\label{eq:cls_loss}
	\mathcal{L}_{cls}(y,\hat{y})= \mathrm{CELoss}(y, \hat{y})
\end{equation}
where $\mathrm{CELoss}(y,p)= -y\log p -(1-y)\log (1-p)$.
For the second goal, we consider the reciprocal roles of the two models and let them supervise \textit{mutually}. In detail, we minimize the cosine similarity between the Pattern and the Fact Preference Map

\begin{equation}\label{eq:dist_loss}
	\mathcal{L}_{cos}= \frac{\mathrm{\mathbf{m_P}}\cdot\mathrm{\mathbf{m_F}}}{\Vert\mathrm{\mathbf{m_P}}\Vert\Vert\mathrm{\mathbf{m_F}}\Vert}
\end{equation}
and the cross-entropy loss under the condition that the input maps for the two models are exchanged and the ground-truth label is reversed
\begin{equation}\label{eq:rev_loss}
	\mathcal{L}_{cls}(y_{rev},\hat{y}^\prime)= \mathrm{CELoss}(y_{rev}, \hat{y}^\prime)
\end{equation}
where $y_{rev}=|1-y|$ and the predictive value $\hat{y}^\prime=\mathrm{MLP}([\mathrm{\mathbf{p}^\prime};\mathrm{\mathbf{f}}^\prime])$. $\mathrm{\mathbf{p}^\prime}$ and $\mathrm{\mathbf{f}}^\prime$ are respectively the output of the pattern-based and the fact-based model with each other's preference map as input. When receiving non-preferred information, the models are expected to be misled and generate non-distinctive features. The total loss of a sample to minimize is
\begin{equation}
	\mathcal{L} = \beta_1\mathcal{L}_{cls}(y,\hat{y}) + \beta_2\mathcal{L}_{cos} + \beta_3\mathcal{L}_{cls}(y_{rev},\hat{y}^\prime)
\end{equation}
where $\beta_1$, $\beta_2$ and $\beta_3$ are trade-off factors in $[0,1]$. We average the loss of samples in each mini-batch before backpropagation. 

\section{Experiments}
We conduct experiments on two datasets to answer the following evaluation questions:

\textbf{EQ1:} Can Pref-FEND improve the performance of fake news detection models with single preference?

\textbf{EQ2:} Can Pref-FEND improve the performance for fake news detection that is integrated by pattern- and fact-based models?

\textbf{EQ3:} How effective are the designed components of Pref-FEND?

\textbf{EQ4:} How different are the Fact and the Pattern Preference Map?

\begin{table}[t]
\centering
\small
\setlength{\tabcolsep}{4pt}
\caption{\label{table:stat} Statistics of the Weibo and the Twitter dataset.}
\begin{tabular}{l|rrr|rrr}
\hline
\multicolumn{1}{c|}{\multirow{2}{*}{\textbf{Number of}}} & \multicolumn{3}{c|}{\textbf{Weibo}} & \multicolumn{3}{c}{\textbf{Twitter}} \\ 
\cline{2-7} 
\multicolumn{1}{c|}{} & Train & Val & Test & Train & Val & Test \\
\hline
Fake News & 1,896 & 632 & 633 & 3,419 & 1,140 & 1,140 \\
Real News & 1,920 & 640 & 641 & 5,406 & 1,802 & 1,802\\ 
\multirow{2}{*}{Total} & 3,816 & 1,272 & 1,274 & 8,825 & 2,942 & 2,942  \\
 & \multicolumn{3}{c|}{(6,362)} & \multicolumn{3}{c}{(14,709)} \\
\hline
\begin{tabular}[c]{@{}l@{}} Relevant Articles\end{tabular} & \multicolumn{3}{c|}{17,849} & \multicolumn{3}{c}{12,419} \\
\hline
\end{tabular}
\end{table}

\subsection{Datasets}
As no existing dataset of fake news detection provides social media posts and relevant articles (as the fact-checking source) simultaneously, we construct two datasets of different languages (Chinese and English) based on the existing data and external sources. The statistics are shown in Table~\ref{table:stat}. The details are as follows:

\noindent\textbf{Weibo Dataset}

\textit{Post.} We utilize the Weibo-20 dataset~\citep{dual-emotion} which contains 6,362 news posts and the ratio of fake and real news posts is roughly 1:1. We keep its original temporal split with a ratio of 6:2:2 for train, validation, and test set.

\textit{Relevant Articles.} We collect fact-checking articles and other relevant articles to construct our fact-checking source. In detail, we use the fact-checking articles crawled in~\citep{sheng-acl} from multiple websites such as \textit{Jiaozhen}\footnote{\url{https://fact.qq.com}}, \textit{Zhuoyaoji}\footnote{\url{http://piyao.sina.cn/}}, and \textit{Baidu Piyao}\footnote{\url{https://author.baidu.com/home?app_id=15060}. \textit{Piyao} means ``refuting false claims''.}. Then, we crawl other relevant articles from Baidu News, with the keywords in the Weibo posts as queries. The keywords are extracted using \texttt{jieba}\footnote{\url{https://github.com/fxsjy/jieba}}. For each query, we obtain at most 30 items and attempt to download full articles using \texttt{Newspaper3k}\footnote{\url{https://newspaper.readthedocs.io/}}. Finally, the de-duplication of all accessible articles lead to an article base containing 17,849 articles.

\noindent\textbf{Twitter Dataset}

 \textit{Post.} We first combine two datasets for detecting previously fact-checked claims released by \citet{shaar2020} and \citet{nguyen2020}, respectively, as they not only provide tweets but also relevant articles from \textit{Snopes}\footnote{\url{https://www.snopes.com/}}. As our task is formulated as a binary classification task, we merge \textit{true}, \textit{mostly-true}, \textit{correct-attribution} into \textit{real}, and \textit{false}, \textit{mostly-false}, \textit{misattributed}, and \textit{legend} into \textit{fake}. The other categories are dropped. As these two datasets are largely imbalanced (1,047 real and 8,992 fake), we utilize PHEME~\citep{allinone} dataset as a supplement, whose annotation files provide some referred news links. For PHEME, we merge \textit{real} and \textit{non-rumor} into \textit{real} and obtain 5,090 real and 638 fake news posts. After pre-processing using \texttt{TexSmart} and dropping failure cases, we obtain 14,709 posts. 
 
 \textit{Relevant Articles.} Because the Twitter dataset has fewer topics than the Weibo dataset, we start from the articles in these datasets to construct the relevant article base. First, we incorporate the fact-checking articles from the datasets released in \citep{shaar2020} and \citep{nguyen2020}, and referred news articles in the PHEME dataset (if accessible). Then, we use their titles (tokenized using NLTK~\citep{nltk}) as queries and search on Google News using \texttt{GNews}\footnote{\url{https://github.com/ranahaani/GNews}}. After post-processing, we obtain an article base containing 12,419 articles. Note that we do not use the existing DeClarE~\citep{declare} and MultiFC~\citep{multifc} datasets which provide both claims (posts) and relevant articles (or webpages) because its claims are normalized and thus with weak patterns of social media posts. We split the train, validation, test set temporally with a ratio of 6:2:2.

\begin{table*}[ht]
\centering
\small
\setlength{\tabcolsep}{3.4pt}
\caption{\label{table:single-results} Performance comparison with pattern- or fact-based models. \textit{w/} $\text{Pref-FEND}_S$ means the model is incorporated as a module of $\text{Pref-FEND}_S$ framework.}
\begin{tabular}{l|>{\columncolor{gray!20}}c>{\columncolor{gray!20}}ccccccc|>{\columncolor{gray!20}}c>{\columncolor{gray!20}}ccccccc}
\hline
\multirow{2}{*}{\textbf{Method}} & \multicolumn{8}{c|}{\textbf{Weibo}} & \multicolumn{8}{c}{\textbf{Twitter}} \\ \cline{2-17} 
 & \multicolumn{1}{c}{Acc.} &  \multicolumn{1}{c}{macF1}  & \multicolumn{1}{c}{P$_{fake}$} & \multicolumn{1}{c}{R$_{fake}$} & \multicolumn{1}{c}{F1$_{fake}$} & \multicolumn{1}{c}{P$_{real}$} & \multicolumn{1}{c}{R$_{real}$} & \multicolumn{1}{c|}{F1$_{real}$} & \multicolumn{1}{c}{Acc.} &  \multicolumn{1}{c}{macF1} & \multicolumn{1}{c}{P$_{fake}$} & \multicolumn{1}{c}{R$_{fake}$} & \multicolumn{1}{c}{F1$_{fake}$} & \multicolumn{1}{c}{P$_{real}$} & \multicolumn{1}{c}{R$_{real}$} & \multicolumn{1}{c}{F1$_{real}$} \\ \hline
\multicolumn{17}{c}{\textbf{Pattern-based}} \\ \hline
Bi-LSTM &  0.667 & 0.660 & 0.626 & 0.820 & 0.710 & 0.744 & 0.516 & 0.610  &  0.767  & 0.732 & 0.753 &0.923 & 0.829 & 0.811 & 0.522 & 0.635  \\ 
\textit{w/}  $\text{Pref-FEND}_S$ & \textbf{0.709} & \textbf{0.709} & 0.696 & 0.735 & 0.715 & 0.723 & 0.683 & 0.702 &  \textbf{0.793} & \textbf{0.788} &0.870 & 0.779 & 0.822 & 0.700 & 0.816 & 0.754   \\ \hline
EANN-Text & 0.692 & 0.690 & 0.860 & 0.785 & 0.717 & 0.739 & 0.601 & 0.663 &  0.770 & 0.725 & 0.742 & 0.960 & 0.837 & 0.881 & 0.472 & 0.614   \\
\textit{w/}  $\text{Pref-FEND}_S$ & \textbf{0.740}& \textbf{0.740} &	0.760&	0.697&	0.727&	0.723&	0.783&	0.752 &  \textbf{0.798} & \textbf{0.788} & 0.837 &0.832 & 0.834 & 0.737 & 0.744 & 0.741    \\ \hline
BERT-Emo & 0.712 & 0.708 &0.667 &	0.839 &	0.743 &	0.787 &	0.587 &	0.672   & 0.794  & 0.762  & 0.769  & 0.950  & 0.850  & 0.873  & 0.550  & 0.675   \\ 
\textit{w/}  $\text{Pref-FEND}_S$ &  \textbf{0.746} & \textbf{0.744} & 0.703 &	0.847 	&0.768 &	0.811 &	0.647 &	0.720   & \textbf{0.804} &	\textbf{0.776} &	0.781 &	0.945 &	0.855 &	0.870 &	0.582 &	0.697  \\ \hline 
\multicolumn{17}{c}{\textbf{Fact-based}} \\ \hline 
DeClarE  & 0.684 & 0.678 & 0.642 & 0.820 & 0.720 & 0.755 & 0.549 & 0.636   & 0.786 & 0.753 & 0.765 & 0.941 & 0.844 & 0.853 & 0.543 & 0.663     \\ 
\textit{w/}  $\text{Pref-FEND}_S$  & \textbf{0.706} & \textbf{0.701} & 0.661 & 0.840 & 0.740 & 0.785 & 0.574 & 0.663 & \textbf{0.798} & \textbf{0.785} & 0.823 & 0.854 & 0.838 & 0.754 & 0.710 & 0.731 \\ \hline
EVIN  & 0.707 &  0.706 & 0.683 & 0.768 & 0.690 & 0.738 & 0.647 & 0.690  & 0.783  & 0.761  & 0.788  & 0.884  & 0.833  & 0.773  & 0.623  & 0.690    \\ 
\textit{w/}  $\text{Pref-FEND}_S$  & \textbf{0.712}& \textbf{0.711} & 0.682 & 0.787 & 0.731 & 0.752 & 0.638 & 0.690   & \textbf{0.795} & \textbf{0.774} & 0.794 & 0.899 & 0.843 & 0.797 & 0.631 & 0.705 \\ \hline
MAC & 0.724 & 0.723  & 0.695 & 0.793 & 0.741 & 0.763 & 0.657 & 0.706     & 0.791  & 0.764  & 0.777  & 0.924  & 0.844  & 0.829  & 0.581  & 0.683   \\
\textit{w/}  $\text{Pref-FEND}_S$  & \textbf{0.749} &  \textbf{0.748} & 0.728 & 0.790 & 0.758 & 0.773 & 0.708 & 0.739   &\textbf{0.804} & \textbf{0.784} & 0.800 & 0.907 & 0.850 & 0.814 & 0.642 & 0.718  \\ \hline
\end{tabular}
\end{table*}

\begin{table*}[ht]
\centering
\small
\setlength{\tabcolsep}{3.2pt}
\caption{\label{table:double-results} Performance comparison with integrated (pattern-and-fact-based) models.}
\begin{tabular}{l|>{\columncolor{gray!20}}c>{\columncolor{gray!20}}ccccccc|>{\columncolor{gray!20}}c>{\columncolor{gray!20}}ccccccc}
\hline
\multirow{2}{*}{\textbf{Method}} & \multicolumn{8}{c|}{\textbf{Weibo}} & \multicolumn{8}{c}{\textbf{Twitter}} \\ \cline{2-17} 
 & \multicolumn{1}{c}{Acc.} &  \multicolumn{1}{c}{macF1}  & \multicolumn{1}{c}{P$_{fake}$} & \multicolumn{1}{c}{R$_{fake}$} & \multicolumn{1}{c}{F1$_{fake}$} & \multicolumn{1}{c}{P$_{real}$} & \multicolumn{1}{c}{R$_{real}$} & \multicolumn{1}{c|}{F1$_{real}$} & \multicolumn{1}{c}{Acc.} &  \multicolumn{1}{c}{macF1} & \multicolumn{1}{c}{P$_{fake}$} & \multicolumn{1}{c}{R$_{fake}$} & \multicolumn{1}{c}{F1$_{fake}$} & \multicolumn{1}{c}{P$_{real}$} & \multicolumn{1}{c}{R$_{real}$} & \multicolumn{1}{c}{F1$_{real}$} \\ \hline
\multicolumn{17}{c}{\textbf{Bi-LSTM (Pattern-based) + DeClarE (Fact-based)}} \\ \hline
Last-layer Fusion & 0.697 &  0.696 & 0.721 & 0.637 & 0.676 & 0.678 & 0.757 & 0.715  & 0.798 & 0.768 & 0.775 & 0.945 & 0.851 & 0.866 & 0.566 & 0.685   \\
Logits Average & 0.692 & 0.685 & 0.646 & 0.840 & 0.730 & 0.776 & 0.544 & 0.640   & 0.784 & 0.750 & 0.762 & 0.943 & 0.843 & 0.855 & 0.534 & 0.657   \\
$\text{Pref-FEND}$  & \textbf{0.714} &  \textbf{0.712} & 0.684 & 0.788 & 0.732 & 0.754 & 0.640 & 0.692  &  \textbf{0.812} & \textbf{0.792} & 0.803 & 0.917 & 0.857 & 0.832 & 0.645 & 0.727 \\ \hline
\multicolumn{17}{c}{\textbf{BERT-Emo (Pattern-based) + MAC (Fact-based)}} \\ \hline
Last-layer Fusion & 0.735 & 0.731 & 0.683 & 0.874 & 0.766 & 0.828 & 0.599 & 0.695  &  0.804 & 0.798 & 0.871 & 0.798 & 0.833 & 0.718 & 0.813 & 0.763  \\
Logits Average &  0.736 & 0.734 & 0.693 & 0.842 & 0.760 & 0.802 & 0.632 & 0.707  & 0.778 & 0.741 & 0.754 & 0.946 & 0.839 & 0.857 & 0.514 & 0.642   \\
$\text{Pref-FEND}$ & \textbf{0.756} & \textbf{0.754} & 0.714 & 0.848 & 0.775 & 0.816 & 0.665 & 0.733  &  \textbf{0.814} & \textbf{0.801} & 0.829 & 0.877 & 0.853 & 0.786 & 0.715 & 0.749  \\ \hline
\end{tabular}
\end{table*}

\subsection{Base Models}
\label{sec:base}
We use six representative text-based models as base models:

\noindent\textbf{Pattern-based Models}
\begin{itemize}[leftmargin=10pt]
	\item \textbf{Bi-LSTM}~\citep{bilstm} is widely used in many existing works of our task for text encoding~\citep{varying-shades, hpa-blstm, hdsf}. We implement a one-layer Bi-LSTM with a maximum sequence length of 100 and a hidden size of 128. We average all the hidden states as representations of posts which are further fed into an MLP for prediction.
	\item \textbf{EANN-Text}~\citep{eann} is a model that tries to distract the fake news detection model from memorizing event-specific features. It uses TextCNN for text representation and adds an auxiliary task of event classification for adversarial learning using gradient reversal layer~\citep{dann}. We re-implement the model according to the public code\footnote{\url{https://github.com/yaqingwang/EANN-KDD18}}. The complete EANN is a multi-modal model but we here use its text-only version. For TextCNN, the number of filters is 20 and the window sizes are $\{1,2,3,4\}$. The labels for the auxiliary event classification task are derived by clustering the training set with K-means where $K=300$. 
	\item \textbf{BERT-Emo}~\citep{dual-emotion} is a model that uses BERT to encode the text and captures the emotion that news publishers express. As we focus on the contents rather than social contexts, we adopt a simplified version where emotions in comments are not considered. We use the author-released code\footnote{\url{https://github.com/RMSnow/WWW2021}}. The maximum sequence length is 150 and the size of embedding vectors is 768.
\end{itemize}

\noindent\textbf{Fact-based Models}
\begin{itemize}[leftmargin=10pt]
	\item \textbf{DeClarE}~\citep{declare} is a model which uses claim-specific attention to focus on salient words in relevant articles. We remove the source embedding which is unavailable in the datasets. We re-implement the model according to the third-party code\footnote{\url{https://github.com/atulkumarin/DeClare/}}. The text encoder is a one-layer Bi-LSTM with the hidden size of 128. 
	\item \textbf{EVIN}~\citep{wulianwei-aaai} is an evidence inference network, which captures the semantic conflicts between the post and relevant articles using the attention mechanism. We re-implement the model according to the paper as no public code is available. The hidden size of one-layer Bi-LSTM is 60. The maximum sequence length is 200.
	\item \textbf{MAC}~\citep{nguyen-eacl} is a hierarchical multi-head attentive network that combines word- and article-level attention. We re-implement according to the author-released code\footnote{\url{https://github.com/nguyenvo09/EACL2021/}}. We use one-layer Bi-LSTM networks with a hidden size of 300 to build MAC. Two multi-head attention modules have 5 and 2 heads, respectively.
\end{itemize}

Note that when base models are used as a module in Pref-FEND, we extract the last-layer feature before the MLP layer.

\subsection{Experimental Setup}
\paragraph{Evaluation Metrics.} We report accuracy (Acc.) and macro F1 score (macF1). For each class, we also report precision, recall, and F1 score, denoted as $P_{cls}$, $R_{cls}$, and ${F1}_{cls}$ where $cls=\{fake, real\}$.

\textit{Implementation Details.}
In Pref-FEND, the number of layers in HetDGCN $L$ is 2.
We perform grid search in a small interval and finally let $\alpha=0.5$, $\beta_1=2$, $\beta_2=1$, and $\beta_3=1$.
For all base models and our Pref-FEND, the initial token embeddings are obtained from pre-trained models in HuggingFace's \texttt{Transformers}\citep{huggingface} (specifically, \texttt{bert-base-chinese} and \texttt{bert-base-uncased}). 
For all fact-based models, the top 5 retrieved articles are considered.
Other hyper-parameters have been described in Section~\ref{sec:base}.
The methods are implemented with PyTorch~\citep{pytorch} and Pytorch Geometric~\citep{pyg}.

\begin{table*}[ht]
\centering
\small
\setlength{\tabcolsep}{2.8pt}
\caption{\label{table:ablation} Ablation study of Pref-FEND.}
\begin{tabular}{l|>{\columncolor{gray!20}}c>{\columncolor{gray!20}}ccccccc|>{\columncolor{gray!20}}c>{\columncolor{gray!20}}ccccccc}
\hline
\multirow{2}{*}{\textbf{Method}} & \multicolumn{8}{c|}{\textbf{Weibo}} & \multicolumn{8}{c}{\textbf{Twitter}} \\ \cline{2-17} 
 & \multicolumn{1}{c}{Acc.} &  \multicolumn{1}{c}{macF1}  & \multicolumn{1}{c}{P$_{fake}$} & \multicolumn{1}{c}{R$_{fake}$} & \multicolumn{1}{c}{F1$_{fake}$} & \multicolumn{1}{c}{P$_{real}$} & \multicolumn{1}{c}{R$_{real}$} & \multicolumn{1}{c|}{F1$_{real}$} & \multicolumn{1}{c}{Acc.} &  \multicolumn{1}{c}{macF1} & \multicolumn{1}{c}{P$_{fake}$} & \multicolumn{1}{c}{R$_{fake}$} & \multicolumn{1}{c}{F1$_{fake}$} & \multicolumn{1}{c}{P$_{real}$} & \multicolumn{1}{c}{R$_{real}$} & \multicolumn{1}{c}{F1$_{real}$} \\ \hline
\multicolumn{17}{c}{\textbf{Bi-LSTM (Pattern-based) + DeClarE (Fact-based)}} \\ \hline
$\text{Pref-FEND}$  & \textbf{0.714} &  \textbf{0.712} & 0.684 & 0.788 & 0.732 & 0.754 & 0.640 & 0.692  &  \textbf{0.812} & \textbf{0.792} & 0.803 & 0.917 & 0.857 & 0.832 & 0.645 & 0.727   \\ \hline
\ \textit{w/ rand init maps} & 0.694 & 0.693 & 0.676 & 0.736 & 0.705 & 0.715 & 0.652 & 0.682   &0.788 & 0.765 & 0.787 & 0.896 & 0.838 & 0.790 & 0.616 & 0.692  \\ \hline 
\ \textit{w/o} $\mathcal{L}_{cos}$ & 0.701 & 0.703 & 0.672 & 0.787 & 0.725 & 0.747 & 0.621 & 0.678   &  0.794 & 0.785 & 0.845 & 0.813 & 0.829 & 0.721 & 0.764 & 0.742   \\
\ \textit{w/o} $\mathcal{L}_{cls}(y_{rev},\hat{y}\prime)$ &  0.703 & 0.702 & 0.710 & 0.679 & 0.694 & 0.696 & 0.725 & 0.710 & 0.792 & 0.764 & 0.775 & 0.932 & 0.846 & 0.842 & 0.571 & 0.681  \\
\ \textit{w/ only} $\mathcal{L}_{cls}(y,\hat{y})$  & 0.700 & 0.702 & 0.672 & 0.782 & 0.723 & 0.743 & 0.622 & 0.677  &  0.789 & 0.747 & 0.752 & 0.979 & 0.851 & 0.936 & 0.490 & 0.643   \\ \hline
\multicolumn{17}{c}{\textbf{BERT-Emo (Pattern-based) + MAC (Fact-based)}} \\ \hline
$\text{Pref-FEND}$  & \textbf{0.756} & \textbf{0.754} & 0.714 & 0.848 & 0.775 & 0.816 & 0.665 & 0.733   &\textbf{0.814} & \textbf{0.801} & 0.829 & 0.877 & 0.853 & 0.786 & 0.715 & 0.749 \\ \hline
\ \textit{w/ rand init maps} &  0.723 & 0.716 & 0.666 & 0.886 & 0.761 & 0.833 & 0.562 & 0.671  &  0.806 & 0.786 & 0.801 & 0.911 & 0.852 & 0.820 & 0.642 & 0.720  \\ \hline 
\ \textit{w/o} $\mathcal{L}_{cos}$ & 0.747 & 0.745 & 0.706 & 0.842 & 0.768 & 0.807 & 0.654 & 0.722  & 0.807 & 0.801 & 0.874 & 0.799 & 0.835 & 0.721 & 0.819 & 0.767   \\
\ \textit{w/o} $\mathcal{L}_{cls}(y_{rev},\hat{y}\prime)$ & 0.745 & 0.740 & 0.690 & 0.883 & 0.775 & 0.841 & 0.608 & 0.706  & 0.808 & 0.789 & 0.806 & 0.903 & 0.852 & 0.811 & 0.657 & 0.726  \\
\ \textit{w/ only} $\mathcal{L}_{cls}(y,\hat{y})$  & 0.741 & 0.735 & 0.682 & 0.896 & 0.775 & 0.851 & 0.588 & 0.696   & 0.792 & 0.787 & 0.869 & 0.778 & 0.821 & 0.699 & 0.815 & 0.752  \\ \hline
\end{tabular}
\end{table*}

\begin{table*}[ht]
\centering
\small
\setlength{\tabcolsep}{2.5pt}
\caption{\label{table:token_analysis} Frequently preferred tokens (separated by ``|'') in the pattern-preferred and the fact-preferred token set.}
\begin{tabular}{p{0.11\textwidth}|l|p{0.74\textwidth}}
\hline
\textbf{Set} & \textbf{Category} & \textbf{Token (Translated into English)} \\ \hline
\multirow{4}{*}{Pattern-preferred} & Punctuation & , | . | ! | :  | ? |  [ | " | | ( | $\ldots$ |  ] |  @ |  < | )  |  \# | $\sim$  |  > |  ; |  / \\
& Negation & not | no | don't  \\
& Pronouns & we | they | you all \\
& Others & find | such | think | may | certainly | so | release | some | as | careful | become | focus | loving heart | but | kind of \\ \hline
\multirow{5}{*}{Fact-preferred} & Evidence-related & claim | video | link | webpage | full text | say | picture | investigation | according to | uncover \\
& Entity-related & China | Beijing | police | place | car | Shanghai | official \\
 & Pronouns & he | its | it | you\\
& Others &  's | done | just | also | and | already | will | wait | go to | do | female | too | want | what | certain | die | pass | death | when | in | second | more | make | suffer | night | society\\
\hline
\end{tabular}
\end{table*}

\subsection{Performance Comparison (\textbf{EQ1} \& \textbf{EQ2})}
\subsubsection{Comparing with Pattern- or Fact-based Methods}
To fairly compare with existing single-preference (i.e., pattern- or fact-based) models, we reduce our framework to a single-model version named $\textbf{Pref-FEND}_S$. In detail, when comparing with a pattern-based model, we remove the fact-based model but preserve the Fact Preference Map for training; and vice versa. From the results in Table~\ref{table:single-results}, we have the following observations:

First, Pref-FEND$_{S}$ successfully improves the performance of all the pattern-based and fact-based models on the two datasets. This verifies our observation that the original base models might be distracted from non-preferred information, which thus limits their generalizability to unseen samples. With the help of Pref-FEND$_{S}$, the base models are more focused during training.

Second, BERT-Emo outperforms Bi-LSTM and EANN-Text. This is as expected because BERT can generate expressive representations and the additional emotion-related features are proved helpful for this task. With the guidance of Pref-FEND$_{S}$, it gains a boost of 3.6 percent points in macro F1 scores on Weibo and a boost of 1.4 percent points on Twitter. This reveals the importance of preference modeling for alleviating the overfitting of specific features.

Third, MAC outperforms DeClarE and EVIN, though they are all based on the attention mechanism. This might be because it effectively uses multi-head attention to capture multi-aspect information. However, some heads might be distracted from the event description in the post, which can be alleviated by our framework.

\subsubsection{Comparing with Integrated (Pattern-and-fact-based) Methods}
We implement the following methods which fuse the information from pattern- and fact-based models:
\begin{itemize}[leftmargin=10pt]
	\item \textbf{Last-layer Fusion} which uses the post as input and concatenates the last-layer features of two models for final prediction;
	\item \textbf{Logits Average} which averages the models' logits (which are in $[0,1]$) for final prediction.
\end{itemize}

We implement these fusion methods and Pref-FEND with two groups of base models, Bi-LSTM+DeClarE and BERT-Emo+MAC. The results are shown in Table~\ref{table:double-results}. Our observations are as follows:

First, Pref-FEND outperforms the two pattern-and-fact-based methods, which validates its effectiveness for integrating pattern- and fact-based models.

Second, comparing with the results in Table~\ref{table:single-results}, Pref-FEND brings further improvements based on the remarkable performance of Pref-FEND$_S$ w.r.t the same base models. For example, on the Weibo dataset, Pref-FEND with Bi-LSTM and DeClarE gains another increase of macro F1 by 0.3 percent points than Pref-FEND$_S$ with Bi-LSTM and 1.1 percent points than Pref-FEND$_S$ with DeClarE. This proves that our framework is applicable to both the single-preference models and the integrated models based on them.

Third, the last-layer fusion does not necessarily perform better than the simple logits average. This indicates that last-layer fusion may be insufficient to align the feature spaces of the pattern- and the fact-based model, which leads to negative fusion effects.

\subsection{Ablation Study (\textbf{EQ3})}
We study the effectiveness of our designed components or strategies based on the Pref-FEND models in Table~\ref{table:double-results}.

\subsubsection{Effectiveness of Model Preference Learning.} Instead of recognizing the entities and stylistic tokens according to the prior knowledge, we randomly initialize preference maps (named as Pref-FEND \textit{w/ rand init maps}). That forces the generation of preference maps to rely only on the supervision of ground-truth labels. The results show that although Pref-FEND \textit{w/ rand init maps} is superior or comparable to the base models on both of the two datasets in terms of accuracy and macro F1, it falls behind the complete Pref-FEND. This proves the effectiveness of our model preference learning, which exploits prior knowledge in a dynamic graph representation learning process.

\subsubsection{Effective of Losses for Differentiating the Preference Maps.} We remove one of the two losses which aim at differentiating the two preference maps, or both. The variants are with the suffixes \textit{w/o} $\mathcal{L}_{cos}$, \textit{w/o} $\mathcal{L}_{cls}(y_{rev},\hat{y}^\prime)$, and \textit{w/ only} $\mathcal{L}_{cls}(y,\hat{y})$, respectively. We see that removing these losses brings performance drops w.r.t. accuracy. The largest drop occurs when removing both the two losses. This indicates that the auxiliary losses are effective and necessary to integrate the two models with different preferences. 

\begin{table*}[t]
\centering
\small
\setlength{\tabcolsep}{2.5pt}
\caption{\label{table:cases} Three examples of fake news posts. \ppppp{Red} represents pattern-preferred tokens and \eeeee{blue} represents fact-preferred tokens. Darker color indicates a higher preference score.}
\begin{tabular}{l|p{0.95\textwidth}}
\hline
\textbf{\#} & \textbf{Post (Translated into English)}\\ \hline
\multirow{2}{*}{1}  & A group of city administration officials in \eeeee{Sishui} , Shandong \p{,} \eeeee{chased} an old man \eeeee{until} all his \eeeee{eggs} were broken on the ground \p{.} The old man \eee{sat} there \ppppp{helplessly} \p{.} The officials ran away \eeeee{after} \eeeee{hitting} \p{.} \e{The white-haired man} should be \eeeee{about} 80 years old \p{,} and he \p{can't} make much money by selling \eeeee{eggs} \p{.} So why be \ppppp{aggressive} \p{?} \eeeee{Is there} \p{no} \e{time} \e{for} the officials \p{to be alone} \p{?} If  the officials \eeeee{only} \p{oppresses} \ee{citizens} \p{,} \p{what's the good} of having these officials \p{?} \p{You} \eeeee{will} \eeeee{be} \p{punished} \ee{sooner or later} \p{for} \p{bullying} the underprivileged \p{.} \\ \cline{2-2}
& \textbf{Ground Truth: Fake} \quad\quad\quad\quad\quad\quad\quad\quad\quad\quad\quad\quad\quad\quad\quad\quad\quad\quad\quad\quad\quad\quad\quad\quad\quad \textbf{Judgment:} Bi-LSTM (Fake), \ \  DeClarE (Real), \ \ Pref-FEND (Fake) \\ \hline
\multirow{2}{*}{2} & \ppppp{[} A student of \eeeee{ZJU} jumping to \eeeee{the West Lake} for a \p{crazy} \eeeee{graduation photo} \eeeee{drowned} \ppppp{]} \e{On June 29} \ppppp{,} Xin \ppppp{(} not his real name \ppppp{)} from \eeeee{ZJU} \eeeee{and} his classmates \eee{went to} \eeeee{the waters} near \eeeee{the scenic spot of} \ppppp{"} \eeeee{Konggu Chuanyin} \ppppp{"} in Gushan , \e{Beili Lake} , \eeeee{West Lake} \eeeee{in Hangzhou} \ppppp{.} Xin \eeeee{asked} his classmates \e{to take} pictures \eeeee{of} his \eeeee{swimming} \ppppp{underwater} \ppppp{.} \eeeee{He} \eeeee{jumped} into \eeeee{the West Lake} \eeeee{from} \eeeee{the side of} \eeeee{Xiling brige on Beishan Road} \ppppp{and} \eeeee{swam} \eeeee{to} \eeeee{the lotus pool} \eeeee{of} Gushan park on the other side \ppppp{.} He \e{drowned} \eeeee{when} \eeeee{swimming} \eee{to} \ppppp{the center of} \eeeee{the lake} \ppppp{.} Recently \ppppp{,} \eeeee{he} has \eeeee{received} a full PhD \ppppp{scholarship} \eeeee{from} a U.S. university .\\ \cline{2-2}
& \textbf{Ground Truth: Fake} \quad\quad\quad\quad\quad\quad\quad\quad\quad\quad\quad\quad\quad\quad\quad\quad\quad\quad\quad\quad\quad\quad\quad\quad\quad \textbf{Judgment:} Bi-LSTM (Real), \ \ DeClarE (Fake), \ \ Pref-FEND (Fake)\\ \hline
\multirow{2}{*}{3} & Is \e{anyone} \eeeee{in Shanghai} interested in \eee{raising a dog} \ppppp{?} \ppppp{No Charge} \ppppp{.} \e{Golden Retriever} \ppppp{,} \eeeee{Poodle} \ppppp{, } \e{Samoyed} \ppppp{, } \ee{and other more kinds} \ppppp{.} \e{There are} dog-killing farms \eeeee{being} \p{destroyed} \ppppp{.} If no one adopts , they \e{will be} \e{euthanized} . Let these \ppppp{little} \ppppp{cute} \ee{lives} \ppppp{accompany with} \eeee{you} \ppppp{.} If you are \ppppp{really} not \e{able to} \eeee{raise them} , please forward this message \ppppp{.} \\ \cline{2-2}
& \textbf{Ground Truth: Fake} \quad\quad\quad\quad\quad\quad\quad\quad\quad\quad\quad\quad\quad\quad\quad\quad\quad\quad\quad\quad\quad\quad\quad\quad\quad \textbf{Judgment:} Bi-LSTM (Real), \ \ DeClarE (Real), \ \ Pref-FEND (Fake)\\ \hline
\end{tabular}
\end{table*}

\subsection{Preference Map Analysis (\textbf{EQ4})}

\subsubsection{Analysis on Most Frequent Token Set.} 
To explore how different the Fact and the Pattern Preference Map are, we analyze the frequently preferred tokens in the Maps. For each post in the validation and test sets of Weibo, we first divide the tokens into a pattern group and a fact group, which indicates this token is scored higher in the Pattern or the Fact Preference Map. Then we extract the top 10 tokens in each group of all the posts and construct two token sets for frequency analysis. The frequent tokens in each set are shown with fine-grained categories in Table~\ref{table:token_analysis}. We see that:

First, in a pattern-preferred token set, punctuations and negation words are important as they express the publishers' tones and emotions. The other frequent tokens are closely related to self-expression, like ``think'', ``may'', and ``kind of''.

Second, in the fact-preferred set, evidence-related tokens that indicate materials and actions (e.g., ``video'', ``webpage'', ``picture'', ``claim'', and ``uncover'') and entity-related tokens (places, positions, etc.) are more focused. Some of the other words do not directly describe an event, but are often around the elements of a news event (e.g., 5W in journalism~\citep{5w}), such as ``already'' and ``when''.

 Third, the focus on pronouns is different between the Pattern and the Fact Preference Map. Plural personal pronouns (``we'', ``they'', and ``you all'') are frequently focused by pattern-based models, while single ones (``he'', ``it'', and ``you'') are preferred by fact-based models. The reason might be that a post with significant fake news patterns often discusses some groups or inspires the audience to take action, while a post with an event description is generally related to specific persons or things.
 
 Our analysis reveals that the learned preference maps are highly correlated to the ideal model preferences and thus effective for the guidance of models' focuses.

\subsubsection{Case study.}  
In Table~\ref{table:cases}, we show three fake news posts that are successfully judged by Pref-FEND with Bi-LSTM and DeClarE.
Case 1 conveys strong signals of emotional patterns, which are preferred by pattern-based models, such as ``helplessly'', and ``aggressive''. Case 2 contains a large number of places and event descriptions, which is friendly to utilize the evidential texts in relevant articles. Due to the different dominant signals, the pattern-based Bi-LSTM judges correctly in Case 1, but fails in Case 2. And the judgments of the fact-based DeClarE are the opposite. However, in Case 3, both of them wrongly judge this post as real. Based on the observation, a pattern-based model can attend to the emotion trigger tokens like ``cute'' and ``really', while a fact-based model can use the place (``Shanghai'') and the dog breed (``Golden Retriever'') to find evidence. Generally, it is unlikely that the two models both fail. We speculate that the failure is led by the negative interference from the non-preferred information. With the help of model preference learning, our Pref-FEND, however, succeed in judging all three posts as fake. These cases demonstrate the necessity of model preference learning and the effectiveness of Pref-FEND.

\section{Conclusion and future work}
We propose the framework Pref-FEND to integrate the pattern-based and fact-based fake news detection models in a preference-aware fashion. The learned preference maps guide the models to focus more on their preferred parts with less interference by the non-preferred parts. Experiments on the two newly constructed datasets show that Pref-FEND outperforms the existing detection models. Further analysis shows that preference learning helps models of different preferences more focused and thus makes both the single-preference and the integrated models better-performing.

How to enhance the interaction between the preference map generation and specific models and how to extend the framework to multi-class and multi-preference scenarios are expected to be explored in the future. The acquisition and exploitation of prior knowledge in this task are also worth studying further to improve overall performance.

\begin{acks}
The authors would like to thank the anonymous reviewers for their valuable comments. This work was supported by the National Key Research and Development Program of China (2017YFC0820604), and the National Natural Science Foundation of China (U1703261).
\end{acks}

\bibliographystyle{ACM-Reference-Format}
 \balance 
\bibliography{sample-base}

\end{document}